\documentclass[10pt,twocolumn,letterpaper]{article}

\usepackage{iccv}
\usepackage{times}
\usepackage{epsfig}
\usepackage{graphicx}
\usepackage{amsmath}
\usepackage{amssymb}

\usepackage[accsupp]{axessibility}
\usepackage[pagebackref=true,breaklinks=true,letterpaper=true,colorlinks,bookmarks=false]{hyperref}
\usepackage[table]{xcolor}
\usepackage[ruled,linesnumbered]{algorithm2e}
\usepackage{comment}
\usepackage{booktabs}
\usepackage{multirow}
\usepackage{caption} 
\usepackage[capitalize]{cleveref}
\crefname{section}{Sec.}{Secs.}
\Crefname{section}{Section}{Sections}
\Crefname{table}{Table}{Tables}
\crefname{table}{Tab.}{Tabs.}

\iccvfinalcopy 

\def\iccvPaperID{5219} 

\ificcvfinal\fi

\begin{document}

\title{AutoSynth: Learning to Generate 3D Training Data\\for Object Point Cloud Registration}

\author{Zheng Dang\textsuperscript{1} and Mathieu Salzmann\textsuperscript{1,2}\\
\textsuperscript{1}CVLab, EPFL, Switzerland\;\;
\textsuperscript{2}ClearSpace, Switzerland\\
{\tt\small \{zheng.dang, mathieu.salzmann\}@epfl.ch}
}

\maketitle
\ificcvfinal\thispagestyle{empty}\fi
\definecolor{mapillarygreen}{RGB}{38,235,179}

\definecolor{blue1}{RGB}{86,174,139}
\definecolor{tabgray}{gray}{.9}

\newcommand{\MS}[1]{{color{red}{{\bf #1}}}}
\newcommand{\ms}[1]{{\color{red}{#1}}}
\newcommand{\ZD}[1]{{\color{blue}{\bf ZD: #1}}}
\newcommand{\zd}[1]{{\color{blue}{#1}}}

\newcommand{\bX}{\mathcal{X}}
\newcommand{\btX}{\tilde{\mathcal{X}}}
\newcommand{\bx}{\mathbf{x}}
\newcommand{\tx}{\tilde{x}}
\newcommand{\bbx}{\bar{\mathbf{x}}}

\newcommand{\bL}{\mathcal{L}}
\newcommand{\bF}{\mathcal{F}}
\newcommand{\bY}{\mathcal{Y}}
\newcommand{\by}{\mathbf{y}}
\newcommand{\bby}{\bar{\mathbf{y}}}
\newcommand{\bfx}{\mathbf{f}_x}
\newcommand{\bfy}{\mathbf{f}_y}
\newcommand{\bthx}{\theta_x}
\newcommand{\bthy}{\theta_y}
\newcommand{\bthxt}{\tilde{\theta}_x}
\newcommand{\bthyt}{\tilde{\theta}_y}
\newcommand{\bK}{\mathbf{K}}
\newcommand{\bT}{\mathcal{T}}
\newcommand{\bt}{\mathbf{t}}
\newcommand{\bS}{\mathcal{S}}
\newcommand{\bbS}{\mathcal{\bar{S}}}
\newcommand{\bP}{\mathcal{P}}
\newcommand{\bbP}{\mathcal{\bar{P}}}
\newcommand{\bM}{\mathcal{M}}
\newcommand{\bQ}{\mathcal{Q}}
\newcommand{\bbM}{\mathcal{\bar{M}}}
\newcommand{\bA}{\mathcal{A}}
\newcommand{\bH}{\mathbf{H}}
\newcommand{\bR}{\mathbf{R}}
\newcommand{\bW}{\mathbf{W}}
\newcommand{\bU}{\mathbf{U}}
\newcommand{\bI}{\mathbf{I}}
\newcommand{\bV}{\mathbf{V}}
\newcommand{\ba}{\mathbf{a}}
\newcommand{\bb}{\mathbf{b}}
\newcommand{\bg}{\mathbf{g}}
\newcommand{\bo}{\mathbf{o}}
\newcommand{\bho}{\mathbf{\hat{o}}}

\newcommand{\hR}{\hat{R}}
\newcommand{\hht}{\hat{\textbf{t}}}
\newcommand{\gR}{R_{gt}}
\newcommand{\gt}{\textbf{t}_{gt}}
\newcommand{\norm}[1]{\left\lVert#1\right\rVert}
\newcommand{\argmin}{\mathop{\mathrm{argmin}}}

\newcommand\largeheight{0.09\columnwidth}
\begin{abstract}
In the current deep learning paradigm, the amount and quality of training data are as critical as the network architecture and its training details. However, collecting, processing, and annotating real data at scale is difficult, expensive, and time-consuming, particularly for tasks such as 3D object registration. 
While synthetic datasets can be created, they require expertise to design and include a limited number of categories.
In this paper, we introduce a new approach called AutoSynth, which automatically generates 3D training data for point cloud registration. Specifically, AutoSynth automatically curates an optimal dataset by exploring a search space encompassing millions of potential datasets with diverse 3D shapes at a low cost.
To achieve this, we generate synthetic 3D datasets by assembling shape primitives, and develop a meta-learning strategy to search for the best training data for 3D registration on real point clouds. For this search to remain tractable, we replace the point cloud registration network with a much smaller surrogate network, leading to a $4056.43$ times speedup. We demonstrate the generality of our approach by implementing it with two different point cloud registration networks, BPNet~\cite{Dang22} and IDAM~\cite{Li20}. Our results on TUD-L~\cite{Hodan18}, LINEMOD~\cite{Hinterstoisser12} and Occluded-LINEMOD~\cite{Brachmann14} evidence that a neural network trained on our searched dataset yields consistently better performance than the same one trained on the widely used ModelNet40 dataset~\cite{Wu15}.
\vspace{-1em}

\end{abstract}
\section{Introduction}

3D point cloud registration, which aims to estimate the relative transformation between two given point clouds, is a traditional computer vision task. With the advent of deep learning, point cloud registration is nowadays commonly tackled with deep networks, achieving impressive results. The main research direction in this area consists of designing new network architectures to improve performance. Here, by contrast, we argue that the quantity and quality of training data have as crucial an impact on the networks' performance as its architecture and training details, and thus advocate data creation as a research goal in itself.

The traditional approach to collecting 3D registration data consists of scanning real objects. This, however, is highly time-consuming and does not scale to the quantity of data commonly expected for deep network training. Generating synthetic data, therefore, comes as a promising alternative. Nevertheless, it requires access to 3D object models, thus often limiting the number of categories, and human expertise to generate realistic data, typically leading to a domain gap w.r.t. real-world point clouds despite best efforts.

In this work, we address this by introducing an approach dubbed AutoSynth, automating the process of curating a 3D dataset. Specifically, we aim for the resulting dataset to act as effective training data for a 3D object registration network that will then be deployed on real-world point clouds. To achieve this, we develop a meta-learning strategy that searches for the optimal dataset over a space encompassing millions of potential datasets, covering a wide diversity of 3D shapes. The search is guided by a target real-world dataset, thus producing data that reduces the domain gap. Our experiments demonstrate that the resulting training dataset yields improved registration performance not only on the target data but on other real-world point clouds.



For this to be possible, we design a very large search space based on the assumption that complex shapes can be created by combining simple primitives. Diverse datasets can then be sampled from this space, and we design an evolutionary algorithm to automatically curate the best training dataset to achieve high performance on the target data. Employing a registration network in the search process, however, would be impractical as even the smallest competitive model would require $1,875$ GPU days on a single RTX8000 for only $1,000$ search steps.
To make the search tractable, we observe that the true quality function, i.e., the accuracy of the registration network of interest, can be replaced with a proxy one, i.e., the reconstruction accuracy of an autoencoder. Specifically, our experiments evidence that, for the same training and testing data, registration accuracy and reconstruction quality follow the same trend, even when using an autoencoder whose architecture is orders of magnitude smaller than that of any registration network able to produce nontrivial results. As such, our approach yields a $4056.43\times$ speedup compared to using a registration network.


We demonstrate the generality of our approach by implementing it with two different point cloud registration networks, BPNet~\cite{Dang22} and IDAM~\cite{Li20}. Our results on TUD-L~\cite{Hodan18}, LINEMOD~\cite{Hinterstoisser12} and Occluded-LINEMOD~\cite{Brachmann14} consistently demonstrate that a neural network trained on our searched dataset achieves better performance than the same one trained on the widely used ModelNet40 dataset~\cite{Wu15}.

Our main contributions can be summarized as follows:
\begin{itemize}
    \item We present AutoSynth, a novel meta-learning-based approach to automatically generate large amounts of 3D training data and curate an optimal dataset for point cloud registration.
    \item We show that the search can be made tractable by leveraging a surrogate network that is $4056.43$ times more efficient than the point cloud registration one.
    \item We evidence that using a single scanned real-object as target dataset during the search yields a training set that leads to good generalization ability.
\end{itemize}
\section{Related Work}
\textbf{Traditional point cloud registration methods.} Point cloud registration aims to estimate the relative pose between two input point sets.
Many algorithms~\cite{Aiger08,Mellado14,Raposo17,Mohamad15,Drost10,Maron16,Rosen19,Jzatt20,Johnson99,Rusu08,Rusu09,Yang19,Le19,Agamennoni16,Hinzmann16,Hahnel02,Fitzgibbon03,Bronstein09,Bronstein08,Gelfand05,Litany12} have contributed to achieving this.
The best-known one probably is Iterative Closest Point (ICP)~\cite{Besl92a}, which has served as basis for many variants, such as Generalized-ICP~\cite{Segal09} and Sparse ICP~\cite{Bouaziz13}, aiming to improve robustness to noise and mismatches. We refer the reader to~\cite{Pomerleau15,Rusinkiewicz01} for a review of ICP-based strategies. The main drawback of ICP-based methods is their requirement for a reasonable initialization to converge to a good solution. As a consequence, recent efforts have been made towards global optimization strategies, leading to algorithms such as Go-ICP~\cite{Yang15}, Super4PCS~\cite{Mellado14}, and Fast Global Registration (FGR)~\cite{Zhou16}. While effective, these methods still suffer from the presence of noise and outliers in the point sets. This is addressed by post-processing strategies, such as that of~\cite{Vidal18}, TEASER~\cite{Yang19}, and TEASER++~\cite{Yang20}. 

\textbf{Learning-based object point cloud registration.} Following the current trend in computer vision, much recent point cloud registration research has focused on a deep learning-based approach. A key requirement to achieve this was the design of deep networks acting on unstructured sets. Deep sets~\cite{Zaheer17} and PointNet~\cite{Qi17} constitute pioneering works in this direction. 
In particular, PointNetLK~\cite{Aoki19} combines the PointNet backbone with the traditional, iterative Lucas-Kanade (LK) algorithm~\cite{Lucas81} so as to form an end-to-end registration network; DCP~\cite{Wang19e} exploits DGCNN~\cite{Wang18b} backbones followed by Transformers~\cite{Vaswani17} to establish 3D-3D correspondences.
While effective, PointNetLK and DCP cannot tackle the partial-to-partial registration scenario. That is, they assume that both point sets are fully observed, during both training and test time. 
PRNet~\cite{Wang19f} and IDAM~\cite{Li20} address this via a deep network designed to extract keypoints from each input set and match these keypoints.
By contrast, RPM-Net~\cite{Yew20} and RGM-Net~\cite{Fu21} build on DCP and adopt a different strategy, replacing the softmax layer with an optimal transport one so as to handle outliers. DeepGMR~\cite{Yuan20} leverages mixtures of Gaussians and formulates registration as the minimization of the KL-divergence between two probability distributions to handle outliers. While the above-mentioned methods were designed to handle point clouds in full 3D, the recent BPNet~\cite{Dang22} was shown to successfully tackle registration from 2.5D measurements, including on real scene datasets, such as TUD-L~\cite{Hodan18}, LM~\cite{Hinterstoisser12} and LMO~\cite{Brachmann14}. Here, we follow an orthogonal direction to these works, and address the task of learning to generate synthetic training data to generalize to real scene test-time observations. We will demonstrate our approach using both BPNet~\cite{Dang22} and IDAM~\cite{Li20}.

\textbf{Learning to generate training data.} Data is essential for the success of learning-based methods, including point cloud registration ones. While much effort has been made to obtain real-world 3D ground truth~\cite{Hodan18,Hinterstoisser12,Brachmann14,Hodan17,Drost17,Kaskman19,Xiang18,Rennie16,Doumanoglou16,Tejani14,Gao22}, synthetic data generation~\cite{Wu15,Xiang16,Liu22a,Wang22,Collins22} has emerged as an effective alternative source for supervision. 
For such synthetic datasets, e.g., ModelNet40~\cite{Wu15}, the creation of each mesh model nonetheless requires human supervision to control its size, position, texture, etc. Hence producing a large amount of synthetic objects remains laborious. 
This raises the question of the feasibility to automatically generate the training data.

In this context, most existing works focus on synthesizing images. For example,
the work of~\cite{Ruiz18}, Meta-Sim~\cite{Kar19}, Meta-Sim2~\cite{Devaranjan20}, and AutoSimulate~\cite{Behl20} learn simulator hyperparameters to maximize the performance of a model on semantic segmentation or object detection.
This is achieved by treating the entire data generation and network training pipeline as a black-box, and using reinforcement learning-based gradient estimation.  
However, these methods still require manually-designed object and scene models as input to the simulator, thus limiting the generated data to a small number of scenes. By contrast, AutoFlow~\cite{Sun21} leverages web images to learn to generate image pairs, thus greatly increasing the data diversity. This, however, does not easily generalize to generating point cloud data.
A few works have nonetheless tackled the problem of generating 3D data using shape primitives. In particular, \cite{Yang18} does so to build training data for a shape-from-shading network that reconstructs object shapes from image sequences; \cite{Yang20a} generates 3D synthetic training data to estimate the surface normals, depth, albedo, and shading maps from a single RGB image. 

Importantly, these techniques rely on the main task network to evaluate the effectiveness of the training data. With the typical growth of state-of-the-art deep network for point cloud registration, this would result in an intractable computational cost. Here, we therefore propose to replace the main task network with a lightweight surrogate network in the searching phase, which we demonstrate to maintain the final performance while requiring three orders of magnitude less computation. 
Note that our approach does not follow the predictor-based strategy commonly used in neural architecture search~\cite{Liu18b,Perez18,Cubuk19}. Specifically, these methods still require training a thousand target models to then train the predictor, which remains too expensive for the computationally-intensive state-of-the-art point cloud registration networks.
Here, instead, we leverage a surrogate network that completely replaces the original one.

\section{Methodology}
\subsection{Problem Formulation}




Our objective is to automatically generate a synthetic 3D dataset $D_{syn}$ such that the main task model (MTM), i.e., a point cloud registration model $\Psi$ in our case, achieves maximum accuracy on the test set when trained on $D_{syn}$ until convergence. The test set is evidently not available during training, and thus we mimic it with a target dataset $D_{tgt}$.

Formally, we express the problem of searching for a synthetic dataset $D_{syn}$ as that of finding a policy $P$, encompassing hyperparameters to generate a 3D dataset, such that $\Psi(w, D_{syn}(P))$ achieves the best performance on $D_{tgt}$.
The set of all policies is referred to as the search space $O$, and we use an evolutionary algorithm to find the best policy $\hat{P}$ that minimizes the evaluation loss
\begin{equation}
    \hat{P} = \underset{P\in O}{\operatorname{argmin}}\;\bL_{eval}(\Psi(w, D_{syn}(P)), D_{tgt}),
    \label{eq:aim}
\end{equation}
where 
$w$ denotes the weights of the MTM trained on $D_{syn}$ until convergence.

\subsection{Search Space}
The search space defines the set of policies that the meta-learning method can explore during training. In other words, it encompasses all possible training datasets, with each policy corresponding to the hyperparameters used to create one 3D dataset.
To generate a dataset, we exploit the observation that complex shapes can be obtained by combining simple primitives~\cite{Clune11,Yang18,Stanley07}, such as cuboids, cones, cylinders, etc.
Following~\cite{Ricci73,Yang18}, we define each shape primitive as an implicit surface function $\bF: \mathbb{R}^3\rightarrow\mathbb{R}$, such that a point $\bx\in \mathbb{R}^3$ on the primitive's surface satisfies $\bF(\bx) = 0$, whereas $\bF(\cdot) < 0$ for interior points and $\bF(\cdot) > 0$ for exterior ones. In other words, $\bF$ encodes a signed distance function. 

Each primitive can then undergo a set of transformations. Specifically, we focus on affine transformations, such as translation, rotation, scaling, shearing, and stretching. For a 3D point $\bx$, this can be expressed as
\begin{equation}
    \bT(\bx) = \alpha T_{rot}T_{shear}T_{stretch}\bx - {\bf t},
\end{equation}
where $\alpha$ is a scaling parameter controlling the overall size of the primitive, ${\bf t}$ is a translation vector, $T_{rot}$ is a rotation matrix, $T_{shear}=S_x S_y S_z$ is a matrix combining shearing operations along the different axes, and $T_{stretch}= A_x A_y A_z$ is a matrix controlling the scale of the primitive along the different axes. Given an existing shape primitive $\bF(\bx)$, the transformed shape can be obtained as $\bF(\bT(\bx))$.

To composite the individual transformed primitives into a complex shape, we utilize logic operators between shapes, as discussed below.
Specifically, to create more distinct shapes from our primitives, we perform truncation with a plane. Let $\bF(\bx)$ denote a transformed primitive, where we neglect the explicit dependency on the transformation $\bT(\cdot)$ for ease of notation. Furthermore, let $\bF_{plane}(\bx)$ denote a plane, defined by a point and a surface normal. The truncation operation can then be expressed as
\begin{equation}
    \bF_{truncation}(\bx) = max(\bF(\bx), \; \bF_{plane}(\bx))\;.
\end{equation}

Given a set of $m$ transformed and truncated primitives with implicit representations $\{\bF_1(\bx), \bF_2(\bx), \dots, \bF_m(\bx)\}$, we combine the shapes using the union operator, which can be expressed as
\begin{equation}
    \bF_{union}(\bx) = \{\bF_1(\bx), \bF_2(\bx), \dots, \bF_m(\bx)\}\;.
\end{equation} 
The final object mesh is generated by merging all vertices and faces of each transformed shape primitive. This operation is substantially faster than mesh union~\cite{Jiang16}, $\bF_{union}(\bx) = min(\bF_1(\bx), \bF_2(\bx), \dots, \bF_m(\bx))$, in practice. The mesh of the primitives can be obtained via the marching cube~\cite{Lorensen87} or simply-defined vertices and faces, and shape generation can be sped up by saving and reusing them.


In this framework, a policy $P$ consists of the 11 parameters corresponding to the above-mentioned 11 operations. Specifically, the operations we search over are rotation~$\{1\}$, translation~~$\{1\}$, overall scale~$\{1\}$, shearing on each axes~$\{3\}$, and stretching on each axes~$\{3\}$, with additional parameters encoding the number of primitives to consider~$\{1\}$ and the truncation plane~$\{1\}$. Each operation also comes with a default range of magnitude. We discretize the range of magnitude into nine values so that we can use a discrete search algorithm to find them. Ultimately, finding the optimal policy $P$ becomes a search problem in a space that contains $9^{11}=31,381,059,609$ possibilities. We refer to this search space as $O$.

Note that the operations described above allow us to form a large search space, which we will show to be effective in practice. However, they are by no means the unique way of defining such a space, and we hope that our work will motivate others to design new search spaces.

\begin{figure}
    \centering
    \includegraphics[width=\linewidth]{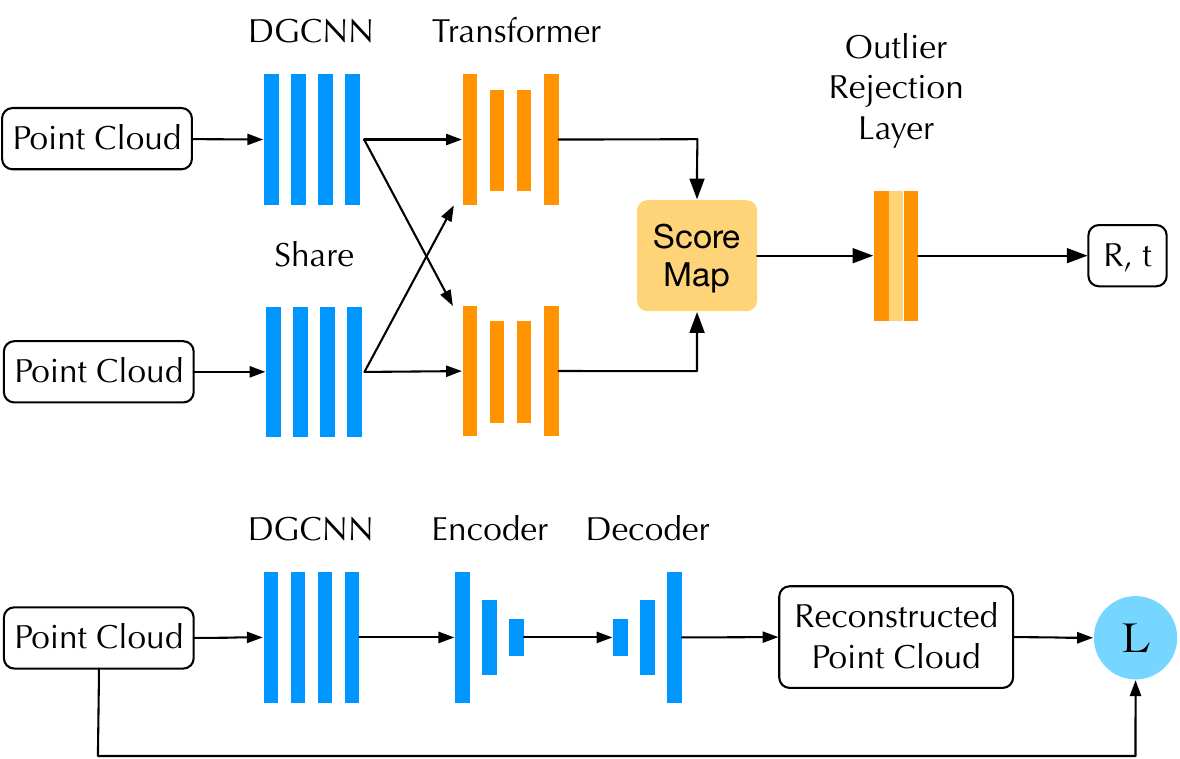}
    \caption{Comparison of the point cloud registration network (top) and the point cloud reconstruction one (bottom)}
    \label{fig:net_comp}
\end{figure}

\subsection{Evolutionary Algorithm}
\label{sec:search}

To automatically search for the optimal policy $P$ in the search space so as to minimize $\bL$ in~\cref{eq:aim}, we employ an evolutionary algorithm with a tournament selection strategy~\cite{Real17}. This algorithm acts as a meta-learner, which iteratively provides policies from which we generate the dataset $D_{syn}(P)$.
The deep network is then trained on the generated dataset $D_{syn}(P)$ and evaluated on $D_{tgt}$ to obtain feedback on its effectiveness. The meta-learner then generates a new policy based on this feedback, which causes the dataset to evolve due to policy changes. This approach allows $D_{tgt}$ to affect the final policy, and if it consists of scanned real-objects, it can help to narrow the domain gap.

Specifically, the evolutionary algorithm starts with an initial population of $k$ policies: $Q = \{P_1, P_2, \dots, P_k | P_i \in O\}$. During each evolutionary step, two individuals $\{P_i, P_j\}$ are chosen from the population $Q$ and their evaluation loss $\{\bL(P_i),\bL(P_j)\}$ are compared, where $\bL(P_i) = \bL_{eval}(\Psi(w, D_{syn}(P_i)), D_{tgt})$.
After each competition, we select the best policy as parent and generate a new policy, $P_{child}$, through mutation. By adding the new $P_{child}$ to the policy pool and removing the worst-performing policy, we ensure that the policy pool remains of the same size and does not shrink.
Specifically, the mutation is performed by randomly choosing one out of the 11 policy hyperparameters from the duplicated best policy and changing this hyperparameter label to another discrete label.
For example, for rotation, assuming a discretization in steps of $\frac{\pi}{8}$, the mutation may change the original label $\frac{\pi}{8}$ to $\frac{3\pi}{8}$.
We then create a new synthetic dataset $D_{syn}(P_{child})$ from the mutated child, and train the network $\Psi$ on it until convergence. At the next evolutionary step, the child has then the possibility of becoming a parent. This process is repeated to evolve policies until a maximum number of trials is reached. We then select the policy with the best evaluation result as the optimal policy $\hat{P}$. The details are given in~\cref{alg:evo}.


\setlength{\textfloatsep}{.5em} 
\begin{algorithm}
    \caption{Evolutionary Policy Search}\label{alg:evo}
    \SetKwInOut{Input}{Input}
    \SetKwInOut{Output}{Output}
    \Input{Search space $O$, population size $k$, max number of trials $M$, target dataset $D_{tgt}$, deep network $\Psi$.}
    \Output{Policy $\hat{P}$ to generate the data that achieves the highest validation performance}
    
    initialize\_population($Q$) by randomly sampling k policies from $O$.
    
    current\_trial\_num := $0$
    
    \While{current\_trial\_num $< M$}{
        1. randomly select two individuals $P_i$ and $P_j$ from $Q$.

        2. train the network $\Psi$ on these two datasets $D_{syn}(P_i)$ and $D_{syn}(P_j)$ until convergence.
        
        3. compare the evaluation loss on the target dataset $D_{tgt}$, and get the best\_individual and the worst\_individual (tournament selection)
        
        4. delete the worst\_individual from $Q$\;
        
        5. mutate the best\_individual, add it to the population $Q$, and train the individual\;
        
        6. current\_trial\_num += 1 \;
    
        }
\end{algorithm}
\vspace{-1.em}
\subsection{Surrogate Task Model}

\label{sec:stm}
The search algorithm described in~\cref{sec:search} requires training a target task model to convergence at every evolutionary trial. Unfortunately, the state-of-the-art point cloud registration networks tend to involve many parameters and expensive layers, such as transformers, as illustrated in the top portion of~\cref{fig:net_comp} for BPNet. As such, searching for the best training data with our search procedure would become prohibitively expensive. For example, using BPNet, one of the smallest registration models, a single trial in the search process would cost $1.875$ GPU days on one Nvidia-V100. Therefore, a standard search process of $1,000$ trials would require $1,875$ GPU days.

To address this, we propose to replace the target task model with a model tackling a surrogate task. For this substitution to make sense, the surrogate model should meet the following conditions: (i) It should take as input the same type of data as the target model, i.e., point clouds; (ii)it should not require any extra annotations; (iii) it should be trainable much more quickly than the task model; (iv) its behavior, i.e., evaluation loss, should follow a similar trend to that of the target model as the training data changes. 
These constraints immediately discard any point registration network, even much reduced versions of existing ones, as we have observed that meaningful registration results can only be obtained with architectures that would be too large for our purpose.
Instead, we propose to make use of a point cloud reconstruction network. This choice was motivated by the observation that, by definition, such a network also operates on point clouds; it does not require any annotations, only the point clouds themselves; it can exploit a much more lightweight architecture, as it does not need to compare two point clouds and thus can be designed without transformer layers.

This leaves the question of evaluation loss behavior. This can be answered from the perspective of multi-task learning literature~\cite{Zamir18}, which has demonstrated that different tasks performed on the same input data often follow similar behavior, i.e., improving one also improves the others.
More pragmatically, we will show in our experiments that the behavior of our point cloud reconstruction network follows that of the registration one as we vary the training set.

\textbf{Surrogate network architecture.}
The architecture of our surrogate network is shown in the bottom portion of~\cref{fig:net_comp}. In essence, it is an autoencoder, relying on the same DGCNN block as the registration network but without any transformer layers. Instead, to prevent the network from directly copying the input point cloud to the output, we project the outputs of the DGCNN to a low-dimensional latent space, and then force the network to reconstruct the whole point cloud from this compressed representation.

Formally, the input to the network is a point set $\bX = \{x_1, \dots, x_v\}$, where $x_i\in \mathbb{R}^3$ represents a 3D point position. We obtain the point set $\bX$ by uniform sampling from the mesh model. The output $\bY$ is the same size point set, representing the reconstructed point positions. The encoder projects each input point cloud into a latent space and the decoder reconstructs the point cloud from the latent representation. We then compute the reconstruction error for $\bY$ using the symmetric Chamfer distance
\begin{equation}
    \bL_{CD} = \frac{1}{2m}(\sum_{x\in\bX}\min_{y\in\bY}\|x - y\|^2_2 + \sum_{y\in\bY}\min_{x\in\bX}\|y - x\|^2_2)\;.
\end{equation}
We train the surrogate network parameter $\theta$ by solving
\begin{equation}
   \theta^* = \argmin_{\theta}\bL_{CD}(\Psi_{surrogate}(\theta, D_{syn})).
\end{equation}
In the searching phase, we therefore also use the symmetric Chamfer distance as fitness score, giving the loss
\begin{equation}
    \bL(P) = \bL_{CD}(\Psi_{surrogate}(\theta^*, D_{syn}(P)), \hat{D}_{tgt})\;.
\end{equation}

Our surrogate task model only needs 15min to converge and only requires $1.42$GB GPU memory. An experiment with $1,000$ trials only takes $0.462$ GPU days on Nvidia-V100 GPU, which is $4056.43$ times more efficient than using the original registration network.
\section{Experiments}
\begin{table*}[!t]
    \centering
    {
    \setlength{\tabcolsep}{2.55mm}{
    \begin{tabular}{l|ccc|ccc|c|cccc}
        \toprule
        & \multicolumn{3}{c|}{Rotation mAP} & \multicolumn{3}{c|}{Translation mAP} & \multicolumn{1}{c|}{ADD} & \multicolumn{4}{c}{BOP Benchmark} \\
        Method & $5^{\circ}$ & $10^{\circ}$ & $20^{\circ}$ & $1cm$ & $2cm$ & $5cm$ & $0.1d$ & VSD & MSSD & MSPD & AR\\
        \midrule
        IDAM-Real                & 0.56 & 0.58 & 0.61 & 0.55 & 0.66 & 0.81 & 0.58 & 0.580 & 0.604 & 0.618 & 0.601\\ 
        BPNet-Real  & \textbf{0.91}& \textbf{0.92}& \textbf{0.93}& \textbf{0.86}& \textbf{0.95}& \textbf{0.99}& \textbf{0.93}& \textbf{0.859} & \textbf{0.914} & \textbf{0.935} & \textbf{0.903}\\
        \midrule
        ICP                 & 0.02 & 0.02 & 0.02 & 0.01 & 0.14 & 0.57 & 0.02 & 0.117 & 0.023 & 0.027 & 0.056 \\
        FGR(FPFH)           & 0.00 & 0.01 & 0.01 & 0.04 & 0.25 & 0.63 & 0.01 & 0.071 & 0.007 & 0.008 & 0.029 \\
        TEASER++(FPFH)      & 0.13 & 0.17 & 0.19 & 0.03 & 0.22 & 0.56 & 0.17 & 0.175 & 0.196 & 0.193 & 0.188 \\
        Super4PCS           & 0.30 & 0.50 & 0.56 & 0.05 & 0.40 & 0.92 & 0.54 & 0.265 & 0.500 & 0.488 & 0.418 \\
        $\star$Vidal-Sensors18 & - & - & - & - & - & - & - & \textbf{0.811} & \textbf{0.910} & \textbf{0.907} & \textbf{0.876} \\
        $\star$Drost        & - & - & - & - & - & - & - & 0.809 & 0.875 & 0.872 & 0.852 \\
        IDAM-MN40           & 0.30 & 0.32 & 0.36 & 0.31 & 0.41 & 0.73 & 0.34 & 0.373 & 0.362 & 0.364 & 0.366 \\ 
        
        IDAM-AutoSynth         & 0.40 & 0.43 & 0.46 & 0.41 & 0.54 & 0.83 & 0.45 & 0.496 & 0.454 & 0.471 & 0.474 \\ 
        BPNet-MN40         & 0.71 & 0.74 & 0.77 & 0.70 & 0.80 & 0.94 & 0.76 & 0.724 & 0.772 & 0.796 & 0.763 \\
        BPNet-AutoSynth        & \textbf{0.78} & \textbf{0.81} & \textbf{0.85} & \textbf{0.77} & \textbf{0.86} & \textbf{0.95} & \textbf{0.84} & 0.777 & 0.845 & 0.867 & 0.829 \\ 
        \bottomrule
    \end{tabular}
    }}
    \caption{Quantitative comparison of registration models trained on AutoSynth and ModelNet40 on the \textbf{TUD-L} real scene dataset. Note that BPNet-Real and IDAM-Real were trained with the TUD-L real scene training sequence, i.e., not in the unseen-object setting. BPNet-MN40 was trained on ModelNet40-full. BPNet-AutoSynth was trained on our AutoSynth generated dataset with the Stanford bunny as target dataset. The results for Vidal-Sensor18~\cite{Vidal18} and Drost (Drost-CVPR10-3D-Edges)~\cite{Drost10} were directly taken from the BOP leaderboard.}
    \label{tab:tudl}
    \vspace{1.em}
\end{table*}

\begin{table*}[!t]
    \setlength{\belowcaptionskip}{-1.em}
    \centering
    {
    \setlength{\tabcolsep}{2.5mm}{
    \begin{tabular}{l|ccc|ccc|c|cccc}
        \toprule
        & \multicolumn{3}{c|}{Rotation mAP} & \multicolumn{3}{c|}{Translation mAP} & \multicolumn{1}{c|}{ADD} & \multicolumn{4}{c}{BOP Benchmark} \\
        Method & $5^{\circ}$ & $10^{\circ}$ & $20^{\circ}$ & $1cm$ & $2cm$ & $5cm$ & $0.1d$ & VSD & MSSD & MSPD & AR\\
        \midrule
        IDAM-Real        & 0.15 & 0.23 & 0.27 & 0.25 & 0.54 & 0.91 & 0.23 & 0.352 & 0.311 & 0.345 & 0.336 \\
        BPNet-Real        & \textbf{0.43} & \textbf{0.59} & \textbf{0.67} & \textbf{0.49} & \textbf{0.83} & \textbf{0.97} & \textbf{0.60} & 0.616 & 0.680 & 0.737 & 0.678 \\
        \midrule
        ICP                 & 0.00 & 0.01 & 0.01 & 0.04 & 0.27 & 0.82 & 0.01 & 0.092 & 0.014 & 0.027 & 0.044 \\
        FGR(FPFH)           & 0.00 & 0.00 & 0.00 & 0.05 & 0.31 & 0.89 & 0.00 & 0.068 & 0.000 & 0.010 & 0.026 \\
        TEASER++(FPFH)      & 0.01 & 0.03 & 0.05 & 0.03 & 0.21 & 0.73 & 0.03 & 0.108 & 0.076 & 0.098 & 0.094 \\
        Super4PCS           & 0.02 & 0.09 & 0.15 & 0.04 & 0.31 & 0.89 & 0.10 & 0.117 & 0.178 & 0.201 & 0.165 \\
        $\star$PPF\_3D\_ICP & - & - & - & - & - & - & - & \textbf{0.719} & \textbf{0.856} & \textbf{0.866} & \textbf{0.814} \\ 
        $\star$Drost & - & - & - & - & - & - & - & 0.678 & 0.786 & 0.789 & 0.751 \\
        IDAM-MN40                        & 0.08 & 0.11 & 0.14 & 0.15 & 0.44 & 0.89 & 0.12 & 0.258 & 0.178 & 0.206 & 0.214 \\
        IDAM-AutoSynth         & 0.21 & 0.29 & 0.33 & 0.28 & 0.60 & 0.91 & 0.29 & 0.420 & 0.359 & 0.398 & 0.392 \\
        BPNet-MN40                       & 0.31 & 0.42 & 0.50 & 0.37 & 0.69 & 0.95 & 0.43 & 0.491 & 0.518 & 0.571 & 0.527 \\
        BPNet-AutoSynth                     & \textbf{0.36} & \textbf{0.49} & \textbf{0.58} & \textbf{0.41} & \textbf{0.74} & \textbf{0.94} & \textbf{0.50} & 0.538 & 0.579 & 0.641 & 0.586 \\
        \bottomrule
    \end{tabular}
    }}
    \caption{Quantitative comparison of registration models trained on AutoSynth and ModelNet40 on the \textbf{LINEMOD} real scene dataset. PPF\_3D\_ICP~\cite{Drost10} and Drost (Drost-CVPR10-3D-Only)~\cite{Drost10} are traditional methods and represent the best depth-only performers from the BOP leaderboard.}
    \label{tab:lm}
\end{table*}

\begin{table*}[!t]
    \centering
    {
    \setlength{\tabcolsep}{2.6mm}{
    \begin{tabular}{l|ccc|ccc|c|cccc}
        \toprule
        & \multicolumn{3}{c|}{Rotation mAP} & \multicolumn{3}{c|}{Translation mAP} & \multicolumn{1}{c|}{ADD} & \multicolumn{4}{c}{BOP Benchmark} \\
        Method & $5^{\circ}$ & $10^{\circ}$ & $20^{\circ}$ & $1cm$ & $2cm$ & $5cm$ & $0.1d$ & VSD & MSSD & MSPD & AR\\
        \midrule
        IDAM-Real        & 0.15 & 0.22 & 0.32 & 0.23 & 0.58 & 0.88 & 0.25 & 0.349 & 0.320 & 0.374 & 0.348 \\
        BPNet-Real          & \textbf{0.31} & \textbf{0.46} & \textbf{0.56} & \textbf{0.37} & \textbf{0.70} & \textbf{0.91} & \textbf{0.47} & 0.478 & 0.542 & 0.612 & 0.544 \\
        \midrule
        ICP                 & 0.01 & 0.01 & 0.01 & 0.07 & 0.36 & 0.85 & 0.01 & 0.085 & 0.014 & 0.032 & 0.044  \\
        FGR(FPFH)           & 0.00 & 0.00 & 0.00 & 0.08 & 0.43 & 0.85 & 0.00 & 0.055 & 0.000 & 0.009 & 0.021  \\
        TEASER++(FPFH)      & 0.01 & 0.02 & 0.05 & 0.04 & 0.26 & 0.77 & 0.02 & 0.096 & 0.060 & 0.093 & 0.083  \\
        Super4PCS           & 0.01 & 0.03 & 0.06 & 0.06 & 0.31 & 0.83 & 0.03 & 0.054 & 0.072 & 0.113 & 0.080 \\
        $\star$Vidal-Sensors18 & - & - & - & - & - & - & - & 0.473 & 0.625 & 0.647 & 0.582 \\
        $\star$PPF\_3D\_ICP & - & - & - & - & - & - & - & \textbf{0.523} & \textbf{0.669} & \textbf{0.716} & \textbf{0.636} \\
        IDAM-MN40                & 0.04 & 0.08 & 0.11 & 0.12 & 0.47 & 0.88 & 0.07 & 0.205 & 0.112 & 0.153 & 0.157 \\
        IDAM-AutoSynth              & 0.14 & 0.21 & 0.26 & 0.23 & 0.57 & 0.88 & 0.20 & 0.316 & 0.272 & 0.322 & 0.303 \\
        BPNet-MN40               & 0.22 & 0.32 & 0.41 & 0.30 & 0.63 & 0.92 & 0.34 & 0.395 & 0.404 & 0.472 & 0.423 \\
        BPNet-AutoSynth             & \textbf{0.25} & \textbf{0.35} & \textbf{0.41} & \textbf{0.34} & \textbf{0.65} & \textbf{0.92} & \textbf{0.37} & 0.410 & 0.429 & 0.501 & 0.447 \\
        \bottomrule
    \end{tabular}
    }}
    \caption{Quantitative comparison of registration models trained on AutoSynth and ModelNet40 on the \textbf{Occluded-LINEMOD} real scene dataset. Vidal-Sensors18~\cite{Vidal18} and PPF\_3D\_ICP~\cite{Drost10} are traditional methods and represent the best depth-only performers from the BOP leaderboard.}
    \label{tab:lmo}
\end{table*}

In this section, we evaluate the effectiveness of our AutoSynth training set search strategy. Below, we first provide implementation details. We then present results on real scenes, and finally analyze different aspects of our approach via ablation studies.

\textbf{Implementation details.}
\label{policy_bunny}
Our complete pipeline consists of two steps: Searching for the best policy using AutoSynth, and training the registration network on the training dataset generated using the best policy.

To generate complex 3D datasets, we utilize a set of shape primitives that includes sphere, cuboid, cone, cylinder, torus, tetrahedron, octahedron, icosahedron, and dodecahedron, as these have shown promising results in our analysis. This set of primitives, however, is not exhaustive and we hope that our results will encourage other researchers to further expand it and propose better alternatives.

In the search process, we build our target dataset $D_{tgt}$ using one scanned real object, i.e., Stanford bunny.
We augment it with random rotations to generate $100$ samples, which constitute $D_{tgt}$. We set the population size to be $32$ and the maximum number of trials to be $1,000$, which we observed to be sufficient to obtain a good policy.
For the reconstruction network, we set the batch size to be $8$ and use the Adam ~\cite{Kingma15} with a learning rate of $0.001$. For each trial in the search phase, we train the reconstruction network for $20,000$ iterations, after which the network has typically converged.
Once the best policy is found, we use it for all the experiments, i.e., we only searched for the policy once.

For BPNet~\cite{Dang22} and IDAM~\cite{Li20}, we use the modified versions of~\cite{Dang22} with Match Normalization. We only replace the training data but keep the same parameter settings as in~\cite{Dang22} to train them to convergence. For the real-scene datasets, we use the provided training sequence.
For ModelNet40~\cite{Wu15}, we use the official training split, which consists of $9,843$ mesh models across $40$ categories. To obtain a source point cloud, we sample points uniformly from a mesh model. For the target point cloud, we generate a depth map from the mesh and with a random camera pose, and sample points from it. For our AutoSynth search process, we only need the source point cloud as input, which also acts as ground truth for the reconstruction network.

Following~\cite{Dang22}, we report the rotation and translation mAP, the ADD, and the BOP benchmark metrics.

\subsection{Results on Real-scene Datasets}
Here, we compare our AutoSynth searched dataset to ModelNet40 by evaluating the performance of the registration models, i.e., BPNet and IDAM, trained on them.
To this end, we evaluate the trained models on three different real-scene datasets, i.e, TUD-L, LM, and LMO. Note that this corresponds to an unseen-object setting, as the training mesh models do not overlap with the test ones.

\textbf{TUD-L dataset.} The results of all methods on TUD-L are summarized in~\cref{tab:tudl}. 
In~\cref{tab:tudl}, the '-Real' model was trained and tested on TUD-L's real scene data, corresponds to the 'seen' object setting. On the other hand, the '-AutoSynth' model, trained on synthetic data and tested on TUD-L's real scene data, represents an 'unseen' object setting. This discrepancy in settings accounts for the observed performance difference. The same principle applies to~\cref{tab:lm,tab:lmo}, which were tested using the LM and LMO datasets.
Furthermore, we also report the results of the top-performing traditional, learning-free, registration methods.

BPNet and IDAM trained on our AutoSynth searched dataset yield significantly better performance than their counterparts trained on ModelNet40. This evidences the superiority of our searched dataset, containing more diverse and complex objects.

Note that the traditional methods based on FPFH features yield poor results. However, `Vidal\-Sensors18' and `CVPR10-3D-Edges', two traditional methods corresponding to the top depth-only performers in the BOP leaderboard, remain more effective than any learning-based method, including ours, in the unseen-object setting.
Nevertheless, we push the limits of what synthetic data can achieve for deep learning-based methods, thus opening the door to future research on learning to generate training data.

The reason why BPNet-Real and IDAM-Real achieve better performance than these models trained on synthetic data is twofold. First, they work in the easier setting where the test object has been observed during training. Second, there remains a domain gap between real-scene depth maps and synthetic ones. While our results show that our AutoSynth approach bridges part of this gap, further reducing it remains a topic for future research.

\begin{table*}[!t]
    \setlength{\belowcaptionskip}{-.5em}
    \centering
    {
    \setlength{\tabcolsep}{3.3mm}{
    \begin{tabular}{l|c|ccc|ccc|c|cccc}
        \toprule
        & \multicolumn{1}{c|}{Method} & \multicolumn{7}{c|}{BPNet} & \multicolumn{1}{c}{AutoEncoder} \\
        \midrule
        \multicolumn{1}{c|}{} & & \multicolumn{3}{c|}{Rotation mAP} & \multicolumn{3}{c|}{Translation mAP} & \multicolumn{1}{c|}{ADD} & \multicolumn{1}{c}{Recontruction} \\
        \multicolumn{1}{c|}{Dataset} & \multicolumn{1}{c|}{Setting}  & $5^{\circ}$ & $10^{\circ}$ & $20^{\circ}$ & $1cm$ & $2cm$ & $5cm$ & $0.1d$ & Chamfer Dist\\
        \midrule
        \multirow{6}{*}{TUD-L} 
        & MN40(01\_per\_cate)         & 0.59 & 0.62 & 0.68 & 0.59 & 0.71 & 0.92 & 0.65 & 22.27 \\ 
        & MN40(05\_per\_cate)         & 0.61 & 0.66 & 0.69 & 0.62 & 0.74 & 0.90 & 0.68 & 17.81 \\ 
        & MN40(10\_per\_cate)         & 0.66 & 0.70 & 0.73 & 0.66 & 0.82 & 0.92 & 0.72 & 9.97 \\ 
        & MN40(50\_per\_cate)         & 0.69 & 0.72 & 0.75 & 0.69 & 0.81 & 0.91 & 0.75 & 6.82 \\  
        & MN40-full                   & 0.71 & 0.74 & 0.77 & 0.70 & 0.80 & 0.94 & 0.76 & 6.65 \\
        & AutoSynth                      & 0.78 & 0.81 & 0.85 & 0.77 & 0.86 & 0.95 & 0.84 & 4.16  \\
        \bottomrule
    \end{tabular}
    }}
    \caption{Comparison of BPNet trained with different datasets and evaluated on the \textbf{TUD-L} real scene dataset. For ModelNet40, we pick 1, 5, 10 and 50 mesh models from each category, corresponding to 40, 200, 400, 2000 mesh models.
    The Chamfer distance is multiplied by $10^3$. Note that, for both registration and reconstruction, increasing the number of training mesh models improves the performance. 
    }
    \label{tab:tendency}
\end{table*}
\textbf{LINEMOD dataset.}
The LINEMOD dataset is more challenging than TUD-L because of the presence of symmetric objects and minor occlusions at the object boundaries. As shown in~\cref{tab:lm}, even Super4PCS fails to yield meaningful results on this dataset. 
Our BPNet-AutoSynth and IDAM-AutoSynth again achieve better performance than BPNet-MN40 and IDAM-MN40. This shows that the dataset searched by our AutoSynth algorithm on the Stanford bunny generalizes well to different real-scene datasets. 
Note that, the IDAM-AutoSynth achieves even better performance than IDAM-Real. This is because the LINEMOD dataset does not provide real depth maps for training data, and we thus used the synthetic ones provided by LINEMOD, which also suffer from a domain gap w.r.t. the real test data. This shows that training on data with more diverse shapes can improve evaluation performance when the domain gap is large.

\textbf{Occluded-LINEMOD dataset.} The Occluded-LINEMOD dataset depicts an even more challenging scenario than LINEMOD by including severe occlusions. As such, as shown in~\cref{tab:lmo}, the results of all the methods deteriorate. Nevertheless, BPNet-AutoSynth and IDAM-AutoSynth still significantly outperform BPNet-MN40 and IDAM-MN40, respectively. This further demonstrates that our searched dataset delivers a consistent performance improvement across different real evaluation datasets and different point cloud registration frameworks.

\subsection{Analysis}
Here, we conduct ablation studies to analyze (i) the behavior similarity of the main and surrogate task networks; (ii) the impact of the target dataset; (iii) the effectiveness of the guidance from the surrogate network; and (iii) the impact of pre-training on the searched data.

\textbf{Behavior of the main and surrogate task networks.}
We conduct experiments to compare the performance of models trained on datasets with different numbers of shapes by adjusting the number of ModelNet40 mesh models used for training. Specifically, we randomly sample $M \in \{1, 5, 10, 50\}$ models per ModelNet40 category. For example, MN40(01\_per\_cate) was built by taking a single mesh model from each category, and thus contains 40 mesh models. For this set of experiments, we use BPNet as our main task registration network.

We summarize the results in~\cref{tab:tendency}, where we also report the reconstruction errors of the surrogate reconstruction network trained on the same data. These results evidence that both tasks, i.e., registration and reconstruction, follow the same trend as the number of training meshes changes. In short, increasing the number of training models improves pose estimation accuracy and lowers reconstruction error. 
Importantly, the results obtained with our AutoSynth searched dataset are the best, confirming the effectiveness of our surrogate task network.

\begin{table}[!t]
    \centering
    {
    \setlength{\tabcolsep}{1.mm}{
    \begin{tabular}{l|ccc|ccc|c}
        \toprule
        & \multicolumn{3}{c|}{Rotation mAP} & \multicolumn{3}{c|}{Translation mAP} & \multicolumn{1}{c}{ADD} \\
        Method & $5^{\circ}$ & $10^{\circ}$ & $20^{\circ}$ & $1cm$ & $2cm$ & $5cm$ & $0.1d$ \\
        \midrule
        MN40(01)   & 0.59 & 0.62 & 0.68 & 0.59 & 0.71 & 0.92 & 0.65 \\
        MN40    & 0.71 & 0.74 & 0.77 & 0.70 & 0.80 & 0.94 & 0.76 \\
        AS(MN40(01))     & 0.76 & 0.80 & 0.84 & 0.77 & 0.86 & 0.93 & 0.82 \\ 
        AS(Real) & \textbf{0.78} & \textbf{0.81} & \textbf{0.85} & \textbf{0.77} & \textbf{0.86} & \textbf{0.95} & \textbf{0.84} \\
        \bottomrule
    \end{tabular}
    }}
    \caption{Results of employing different datasets as the target dataset with BPNet as backbone.
    MN40(01) stands for MN40(01\_per\_cate); AS stands for AutoSynth.}
    \label{tab:target_dataset}
    \vspace{1.5em}
\end{table}
\textbf{Impact of the target dataset $D_{tgt}$.}
To assess the influence of $D_{tgt}$ on the search, we compare the use of a scanned real-object with an MN40(01\_per\_cate) dataset, using BPNet as backbone.
Our framework leverages a feedback mechanism to learn from $D_{tgt}$, which helps to narrow the reality gap when using scanned real-objects. The results presented in~\cref{tab:target_dataset} show that AutoSynth(Real) outperforms AutoSynth(MN40(01\_per\_cate)), which confirms our claim.
Note that BPNet trained using MN40(01\_per\_cate) as $D_{syn}$ yields better results than the one trained on it directly. This is due to the fact that the 3D dataset evolved from MN40(01\_per\_cate) contains more distinct shapes than it, resulting in better performance.

\textbf{Effectiveness of the guidance from the surrogate network.}
\begin{table}[!t]
    \centering
    {
    \setlength{\tabcolsep}{1.15mm}{
    \begin{tabular}{l|ccc|ccc|c}
        \toprule
        & \multicolumn{3}{c|}{Rotation mAP} & \multicolumn{3}{c|}{Translation mAP} & \multicolumn{1}{c}{ADD} \\
        Method & $5^{\circ}$ & $10^{\circ}$ & $20^{\circ}$ & $1cm$ & $2cm$ & $5cm$ & $0.1d$ \\
        \midrule
        full-range    & 0.66 & 0.68 & 0.70 & 0.64 & 0.75 & 0.90 & 0.69\\ 
        no-feedback & 0.61 & 0.65 & 0.69 & 0.62 & 0.74 & 0.89 & 0.68\\ 
        Surrogate net        & \textbf{0.78} & \textbf{0.81} & \textbf{0.85} & \textbf{0.77} & \textbf{0.86} & \textbf{0.95} & \textbf{0.84} \\
        \bottomrule
    \end{tabular}
    }}
    \caption{Effectiveness of the feedback mechanism. Using our surrogate reconstruction network to guide the search clearly outperforms both selecting a random policy and using the full range policy.}
    \label{tab:ablation}
    \vspace{2em}
\end{table}
Here, we evaluate the effectiveness of the surrogate network $\Psi_{surrogate}$ at guiding the search towards the best policy by comparing it with two alternatives that offer no guidance: (i) A no-feedback strategy corresponding to randomly picking a policy from the search space; (ii) a full-range policy consisting of randomly sampling using the largest possible range of transformations during training. The comparison in~\cref{tab:ablation} on the TUD-L dataset testing sequence and with BPNet as registration network clearly shows the benefits of the surrogate network for the search.

\begin{table}[t]
    \centering
    {
    \setlength{\tabcolsep}{1.35mm}{
    \begin{tabular}{l|ccc|ccc|c}
        \toprule
        & \multicolumn{3}{c|}{Rotation mAP} & \multicolumn{3}{c|}{Translation mAP} & \multicolumn{1}{c}{ADD} \\
        Method & $5^{\circ}$ & $10^{\circ}$ & $20^{\circ}$ & $1cm$ & $2cm$ & $5cm$ & $0.1d$\\
        \midrule
        Real             &  0.91 &  0.92 &  0.93 &  0.86 &  0.95 &  0.99 &  0.93 \\
        AutoSynth        &  0.78 &  0.81 &  0.85 &  0.77 &  0.86 &  0.95 &  0.84  \\ 
        Pretrain            &  \textbf{0.94} &  \textbf{0.95} &  \textbf{0.96} &  \textbf{0.90} &  \textbf{0.97} &  \textbf{1.00} &  \textbf{0.96} \\ 
        \bottomrule
    \end{tabular}
    }}
    \caption{BPNet trained on TUD-L vs AutoSynth vs AutoSynth pre-trained followed by TUD-L fine-tuning.}
    \label{tab:rebuttal_tudl}
    \vspace{1.em}
\end{table}

\textbf{Impact of pre-training on the searched data.}
To evaluate the use of our approach as a pre-training strategy, we pre-train the network on the AutoSynth-searched data and fine-tune it on the TUD-L training set. As shown in  Tab.~\ref{tab:rebuttal_tudl}, this lets us reach a new SOTA performance (\textbf{0.94} in R$5^{\circ}$ mAP), showing the effectiveness of our AutoSynth dataset. 

\section{Conclusion}
We have introduced a novel algorithm to automatically generate large amounts of 3D training dataset and curate the optimal one from the millions of options. To this end, we have proposed to use a surrogate reconstruction network while searching for a data generation policy, thus accelerating the search by $4056.43$ times. We have evidenced the generality of our approach by evaluating it with two different point cloud registration methods, BPNet and IDAM. Our experiments on real-scene datasets have evidenced that a network trained on our searched dataset consistently outperforms the same model trained on the widely used ModelNet40 dataset. As shown by our results, however, there remains a gap between our searched dataset and real scans. In the future, we will study how to further bridge this gap by improving the realism of the synthesized data.
\section{Acknowledgements}
Zheng Dang would like to thank H. Chen for the highly-valuable discussions and for her encouragement. This work was funded in part by the Swiss Innovation Agency (Innosuisse).

{\small
\bibliographystyle{ieee_fullname}
\bibliography{bibtex/string,bibtex/vision}

\begin{thebibliography}{10}\itemsep=-1pt

\bibitem{Agamennoni16}
Gabriel Agamennoni, Simone Fontana, Roland~Y Siegwart, and Domenico~G Sorrenti.
\newblock Point clouds registration with probabilistic data association.
\newblock In {\em 2016 IEEE/RSJ International Conference on Intelligent Robots
  and Systems (IROS)}, pages 4092--4098. IEEE, 2016.

\bibitem{Aiger08}
Dror Aiger, Niloy~J Mitra, and Daniel Cohen-Or.
\newblock 4-points congruent sets for robust pairwise surface registration.
\newblock In {\em ACM SIGGRAPH 2008 papers}, pages 1--10, 2008.

\bibitem{Aoki19}
Yasuhiro Aoki, Hunter Goforth, Rangaprasad~Arun Srivatsan, and Simon Lucey.
\newblock Pointnetlk: Robust \& efficient point cloud registration using
  pointnet.
\newblock In {\em Conference on Computer Vision and Pattern Recognition}, pages
  7163--7172, Long Beach, California, 2019.

\bibitem{Behl20}
Harkirat~Singh Behl, Atilim~G{\"u}ne{\c{s}} Baydin, Ran Gal, Philip~HS Torr,
  and Vibhav Vineet.
\newblock Autosimulate:(quickly) learning synthetic data generation.
\newblock In {\em European Conference on Computer Vision}, pages 255--271.
  Springer, 2020.

\bibitem{Besl92a}
P. Besl and N. Mckay.
\newblock A method for registration of 3d shapes.
\newblock {\em IEEE Transactions on Pattern Analysis and Machine Intelligence},
  14(2):239--256, February 1992.

\bibitem{Bouaziz13}
Sofien Bouaziz, Andrea Tagliasacchi, and Mark Pauly.
\newblock Sparse iterative closest point.
\newblock In {\em Computer graphics forum}, volume~32, pages 113--123, Hoboken,
  New Jersey, 2013. Wiley Online Library.

\bibitem{Brachmann14}
Eric Brachmann, Alexander Krull, Frank Michel, Stefan Gumhold, Jamie Shotton,
  and Carsten Rother.
\newblock Learning 6d object pose estimation using 3d object coordinates.
\newblock In {\em European conference on computer vision}, pages 536--551,
  Zürich, Switzerland, 2014. Springer.

\bibitem{Bronstein08}
Alexander~M Bronstein and Michael~M Bronstein.
\newblock Regularized partial matching of rigid shapes.
\newblock In {\em European Conference on Computer Vision}, pages 143--154.
  Springer, 2008.

\bibitem{Bronstein09}
Alexander~M Bronstein, Michael~M Bronstein, Alfred~M Bruckstein, and Ron
  Kimmel.
\newblock Partial similarity of objects, or how to compare a centaur to a
  horse.
\newblock {\em International Journal of Computer Vision}, 84(2):163--183, 2009.

\bibitem{Clune11}
Jeff Clune and Hod Lipson.
\newblock Evolving 3d objects with a generative encoding inspired by
  developmental biology.
\newblock {\em ACM SIGEVOlution}, 5(4):2--12, 2011.

\bibitem{Collins22}
Jasmine Collins, Shubham Goel, Kenan Deng, Achleshwar Luthra, Leon Xu, Erhan
  Gundogdu, Xi Zhang, Tomas F~Yago Vicente, Thomas Dideriksen, Himanshu Arora,
  et~al.
\newblock Abo: Dataset and benchmarks for real-world 3d object understanding.
\newblock In {\em Proceedings of the IEEE/CVF Conference on Computer Vision and
  Pattern Recognition}, pages 21126--21136, 2022.

\bibitem{Cubuk19}
Ekin~D Cubuk, Barret Zoph, Dandelion Mane, Vijay Vasudevan, and Quoc~V Le.
\newblock Autoaugment: Learning augmentation strategies from data.
\newblock In {\em Proceedings of the IEEE/CVF Conference on Computer Vision and
  Pattern Recognition}, pages 113--123, 2019.

\bibitem{Dang22}
Zheng Dang, Wang Lizhou, Guo Yu, and Mathieu Salzmann.
\newblock Learning-based point cloud registration for 6d object pose estimation
  in the real world.
\newblock In {\em European Conference on Computer Vision}, 2022.

\bibitem{Devaranjan20}
Jeevan Devaranjan, Amlan Kar, and Sanja Fidler.
\newblock Meta-sim2: Unsupervised learning of scene structure for synthetic
  data generation.
\newblock In {\em European Conference on Computer Vision}, pages 715--733.
  Springer, 2020.

\bibitem{Doumanoglou16}
Andreas Doumanoglou, Rigas Kouskouridas, Sotiris Malassiotis, and Tae-Kyun Kim.
\newblock Recovering 6d object pose and predicting next-best-view in the crowd.
\newblock In {\em Proceedings of the IEEE conference on computer vision and
  pattern recognition}, pages 3583--3592, Las Vegas, Nevada, 2016.

\bibitem{Drost17}
Bertram Drost, Markus Ulrich, Paul Bergmann, Philipp Hartinger, and Carsten
  Steger.
\newblock Introducing mvtec itodd-a dataset for 3d object recognition in
  industry.
\newblock In {\em Proceedings of the IEEE International Conference on Computer
  Vision Workshops}, pages 2200--2208, Venice, Italy, 2017.

\bibitem{Drost10}
Bertram Drost, Markus Ulrich, Nassir Navab, and Slobodan Ilic.
\newblock Model globally, match locally: Efficient and robust 3d object
  recognition.
\newblock In {\em 2010 IEEE computer society conference on computer vision and
  pattern recognition}, pages 998--1005, 2010.

\bibitem{Fitzgibbon03}
Andrew~W Fitzgibbon.
\newblock Robust registration of 2d and 3d point sets.
\newblock {\em Image and vision computing}, 21(13-14):1145--1153, 2003.

\bibitem{Fu21}
Kexue Fu, Shaolei Liu, Xiaoyuan Luo, and Manning Wang.
\newblock Robust point cloud registration framework based on deep graph
  matching.
\newblock In {\em Proceedings of the IEEE/CVF Conference on Computer Vision and
  Pattern Recognition}, pages 8893--8902, 2021.

\bibitem{Gao22}
Ruohan Gao, Zilin Si, Yen-Yu Chang, Samuel Clarke, Jeannette Bohg, Li Fei-Fei,
  Wenzhen Yuan, and Jiajun Wu.
\newblock Objectfolder 2.0: A multisensory object dataset for sim2real
  transfer.
\newblock In {\em Proceedings of the IEEE/CVF Conference on Computer Vision and
  Pattern Recognition}, pages 10598--10608, 2022.

\bibitem{Gelfand05}
Natasha Gelfand, Niloy~J Mitra, Leonidas~J Guibas, and Helmut Pottmann.
\newblock Robust global registration.
\newblock In {\em Symposium on geometry processing}, page~5. Vienna, Austria,
  2005.

\bibitem{Hahnel02}
Dirk H{\"a}hnel and Wolfram Burgard.
\newblock Probabilistic matching for 3d scan registration.
\newblock In {\em Proc. of the VDI-Conference Robotik}, volume 2002. Citeseer,
  2002.

\bibitem{Hinterstoisser12}
Stefan Hinterstoisser, Vincent Lepetit, Slobodan Ilic, Stefan Holzer, Gary
  Bradski, Kurt Konolige, and Nassir Navab.
\newblock Model based training, detection and pose estimation of texture-less
  3d objects in heavily cluttered scenes.
\newblock In {\em Asian Conference on Computer Vision}, pages 548--562,
  Daejeon, 2012.

\bibitem{Hinzmann16}
Timo Hinzmann, Thomas Stastny, Gianpaolo Conte, Patrick Doherty, Piotr Rudol,
  Marius Wzorek, Enric Galceran, Roland Siegwart, and Igor Gilitschenski.
\newblock Collaborative 3d reconstruction using heterogeneous uavs: System and
  experiments.
\newblock In {\em International Symposium on Experimental Robotics}, pages
  43--56. Springer, 2016.

\bibitem{Hodan17}
Tom{\'a}{\v{s}} Hodan, Pavel Haluza, {\v{S}}tep{\'a}n Obdr{\v{z}}{\'a}lek, Jiri
  Matas, Manolis Lourakis, and Xenophon Zabulis.
\newblock T-less: An rgb-d dataset for 6d pose estimation of texture-less
  objects.
\newblock In {\em 2017 IEEE Winter Conference on Applications of Computer
  Vision (WACV)}, pages 880--888, Santa Rosa, CA, USA, 2017. IEEE.

\bibitem{Hodan18}
Tomas Hodan, Frank Michel, Eric Brachmann, Wadim Kehl, Anders GlentBuch, Dirk
  Kraft, Bertram Drost, Joel Vidal, Stephan Ihrke, Xenophon Zabulis, et~al.
\newblock {Bop: Benchmark for 6D Object Pose Estimation}.
\newblock In {\em European Conference on Computer Vision}, pages 19--34,
  Munich, Germany, 2018.

\bibitem{Jzatt20}
Gregory Izatt, Hongkai Dai, and Russ Tedrake.
\newblock Globally optimal object pose estimation in point clouds with
  mixed-integer programming.
\newblock In {\em Robotics Research}, pages 695--710, Ventura, CA, 2020.
  Springer.

\bibitem{Jiang16}
Xiaotong Jiang, Qingjin Peng, Xiaosheng Cheng, Ning Dai, Cheng Cheng, and Dawei
  Li.
\newblock Efficient booleans algorithms for triangulated meshes of geometric
  modeling.
\newblock {\em Computer-Aided Design and Applications}, 13(4):419--430, 2016.

\bibitem{Johnson99}
Andrew~E. Johnson and Martial Hebert.
\newblock Using spin images for efficient object recognition in cluttered 3d
  scenes.
\newblock {\em TPAMI}, 21(5):433--449, 1999.

\bibitem{Kar19}
Amlan Kar, Aayush Prakash, Ming-Yu Liu, Eric Cameracci, Justin Yuan, Matt
  Rusiniak, David Acuna, Antonio Torralba, and Sanja Fidler.
\newblock Meta-sim: Learning to generate synthetic datasets.
\newblock In {\em Conference on Computer Vision and Pattern Recognition}, pages
  4551--4560, 2019.

\bibitem{Kaskman19}
Roman Kaskman, Sergey Zakharov, Ivan Shugurov, and Slobodan Ilic.
\newblock Homebreweddb: Rgb-d dataset for 6d pose estimation of 3d objects.
\newblock In {\em Proceedings of the IEEE International Conference on Computer
  Vision Workshops}, pages 0--0, Seoul, Korea, 2019.

\bibitem{Kingma15}
Diederik~P Kingma and Jimmy Ba.
\newblock Adam: A method for stochastic optimization.
\newblock In {\em International Conference on Learning Representations}, San
  Diego, CA, USA, 2015.

\bibitem{Le19}
Huu~M Le, Thanh-Toan Do, Tuan Hoang, and Ngai-Man Cheung.
\newblock Sdrsac: Semidefinite-based randomized approach for robust point cloud
  registration without correspondences.
\newblock In {\em Proceedings of the IEEE/CVF Conference on Computer Vision and
  Pattern Recognition}, pages 124--133, 2019.

\bibitem{Li20}
Jiahao Li, Changhao Zhang, Ziyao Xu, Hangning Zhou, and Chi Zhang.
\newblock Iterative distance-aware similarity matrix convolution with
  mutual-supervised point elimination for efficient point cloud registration.
\newblock In {\em ECCV 2020: 16th European Conference, Glasgow, UK, August
  23--28, 2020, Proceedings, Part XXIV 16}, pages 378--394. Springer, 2020.

\bibitem{Litany12}
Or Litany, Alexander~M Bronstein, and Michael~M Bronstein.
\newblock Putting the pieces together: Regularized multi-part shape matching.
\newblock In {\em European Conference on Computer Vision}, pages 1--11.
  Springer, 2012.

\bibitem{Liu18b}
Chenxi Liu, Barret Zoph, Maxim Neumann, Jonathon Shlens, Wei Hua, Li-Jia Li, Li
  Fei-Fei, Alan Yuille, Jonathan Huang, and Kevin Murphy.
\newblock Progressive neural architecture search.
\newblock In {\em European Conference on Computer Vision}, pages 19--34, 2018.

\bibitem{Liu22a}
Liu Liu, Wenqiang Xu, Haoyuan Fu, Sucheng Qian, Qiaojun Yu, Yang Han, and Cewu
  Lu.
\newblock Akb-48: A real-world articulated object knowledge base.
\newblock In {\em Proceedings of the IEEE/CVF Conference on Computer Vision and
  Pattern Recognition}, pages 14809--14818, 2022.

\bibitem{Lorensen87}
William~E Lorensen and Harvey~E Cline.
\newblock Marching cubes: A high resolution 3d surface construction algorithm.
\newblock {\em ACM siggraph computer graphics}, 21(4):163--169, 1987.

\bibitem{Lucas81}
Bruce~D Lucas, Takeo Kanade, et~al.
\newblock An iterative image registration technique with an application to
  stereo vision.
\newblock In {\em International Joint Conference on Artificial Intelligence},
  Vancouver, British Columb, 1981.

\bibitem{Maron16}
Haggai Maron, Nadav Dym, Itay Kezurer, Shahar Kovalsky, and Yaron Lipman.
\newblock Point registration via efficient convex relaxation.
\newblock {\em ACM Transactions on Graphics (TOG)}, 35(4):1--12, 2016.

\bibitem{Mellado14}
Nicolas Mellado, Dror Aiger, and Niloy~J Mitra.
\newblock Super 4pcs fast global pointcloud registration via smart indexing.
\newblock In {\em Computer graphics forum}, volume~33, pages 205--215. Wiley
  Online Library, 2014.

\bibitem{Mohamad15}
Mustafa Mohamad, Mirza~Tahir Ahmed, David Rappaport, and Michael Greenspan.
\newblock Super generalized 4pcs for 3d registration.
\newblock In {\em 2015 International Conference on 3D Vision}, pages 598--606.
  IEEE, 2015.

\bibitem{Perez18}
Juan-Manuel Perez-Rua, Moez Baccouche, and Stephane Pateux.
\newblock Efficient progressive neural architecture search.
\newblock {\em arXiv preprint arXiv:1808.00391}, 2018.

\bibitem{Pomerleau15}
Fran{\c{c}}ois Pomerleau, Francis Colas, Roland Siegwart, et~al.
\newblock A review of point cloud registration algorithms for mobile robotics.
\newblock {\em Foundations and Trends{\textregistered} in Robotics},
  4(1):1--104, 2015.

\bibitem{Qi17}
C.R. Qi, H. Su, K. Mo, and L.J. Guibas.
\newblock Pointnet: Deep learning on point sets for 3d classification and
  segmentation.
\newblock In {\em Conference on Computer Vision and Pattern Recognition},
  Honolulu, Hawaii, 2017.

\bibitem{Raposo17}
Carolina Raposo and Joao~P Barreto.
\newblock Using 2 point+ normal sets for fast registration of point clouds with
  small overlap.
\newblock In {\em 2017 IEEE International Conference on Robotics and Automation
  (ICRA)}, pages 5652--5658. IEEE, 2017.

\bibitem{Real17}
Esteban Real, Sherry Moore, Andrew Selle, Saurabh Saxena, Yutaka~Leon Suematsu,
  Jie Tan, Quoc~V Le, and Alexey Kurakin.
\newblock Large-scale evolution of image classifiers.
\newblock In {\em International Conference on Machine Learning}, pages
  2902--2911. PMLR, 2017.

\bibitem{Rennie16}
Colin Rennie, Rahul Shome, Kostas~E Bekris, and Alberto~F De~Souza.
\newblock A dataset for improved rgbd-based object detection and pose
  estimation for warehouse pick-and-place.
\newblock {\em IEEE Robotics and Automation Letters}, 1(2):1179--1185, 2016.

\bibitem{Ricci73}
Antonio Ricci.
\newblock A constructive geometry for computer graphics.
\newblock {\em The Computer Journal}, 16(2):157--160, 1973.

\bibitem{Rosen19}
David~M Rosen, Luca Carlone, Afonso~S Bandeira, and John~J Leonard.
\newblock Se-sync: A certifiably correct algorithm for synchronization over the
  special euclidean group.
\newblock {\em The International Journal of Robotics Research},
  38(2-3):95--125, 2019.

\bibitem{Ruiz18}
Nataniel Ruiz, Samuel Schulter, and Manmohan Chandraker.
\newblock Learning to simulate.
\newblock In {\em International Conference on Learning Representations}, 2019.

\bibitem{Rusinkiewicz01}
Szymon Rusinkiewicz and Marc Levoy.
\newblock Efficient variants of the icp algorithm.
\newblock In {\em Proceedings Third International Conference on 3-D Digital
  Imaging and Modeling}, pages 145--152, Quebec City, Canada, 2001. IEEE.

\bibitem{Rusu09}
Radu~Bogdan Rusu, Nico Blodow, and Michael Beetz.
\newblock Fast point feature histograms (fpfh) for 3d registration.
\newblock In {\em International Conference on Robotics and Automation}, pages
  3212--3217, Kobe, Japan, 2009. IEEE.

\bibitem{Rusu08}
Radu~Bogdan Rusu, Nico Blodow, Zoltan~Csaba Marton, and Michael Beetz.
\newblock Aligning point cloud views using persistent feature histograms.
\newblock In {\em International Conference on Intelligent Robots and Systems},
  pages 3384--3391, Nice, France, 2008. IEEE.

\bibitem{Segal09}
Aleksandr Segal, Dirk Haehnel, and Sebastian Thrun.
\newblock Generalized-icp.
\newblock In {\em In Robotics: Science and Systems}, Cambridge, 2009.

\bibitem{Stanley07}
Kenneth~O Stanley.
\newblock Compositional pattern producing networks: A novel abstraction of
  development.
\newblock {\em Genetic programming and evolvable machines}, 8(2):131--162,
  2007.

\bibitem{Sun21}
Deqing Sun, Daniel Vlasic, Charles Herrmann, Varun Jampani, Michael Krainin,
  Huiwen Chang, Ramin Zabih, William~T Freeman, and Ce Liu.
\newblock Autoflow: Learning a better training set for optical flow.
\newblock In {\em Proceedings of the IEEE/CVF Conference on Computer Vision and
  Pattern Recognition}, pages 10093--10102, 2021.

\bibitem{Tejani14}
Alykhan Tejani, Danhang Tang, Rigas Kouskouridas, and Tae-Kyun Kim.
\newblock Latent-class hough forests for 3d object detection and pose
  estimation.
\newblock In {\em European Conference on Computer Vision}, pages 462--477,
  Zürich, Switzerland, 2014. Springer.

\bibitem{Vaswani17}
Ashish Vaswani, Noam Shazeer, Niki Parmar, Jakob Uszkoreit, Llion Jones,
  Aidan~N Gomez, {\L}ukasz Kaiser, and Illia Polosukhin.
\newblock Attention is all you need.
\newblock In {\em Advances in Neural Information Processing Systems}, pages
  5998--6008, Long Beach, California, United States, 2017.

\bibitem{Vidal18}
Joel Vidal, Chyi-Yeu Lin, Xavier Llad{\'o}, and Robert Mart{\'\i}.
\newblock A method for 6d pose estimation of free-form rigid objects using
  point pair features on range data.
\newblock {\em Sensors}, 18(8):2678, 2018.

\bibitem{Wang22}
Pengyuan Wang, HyunJun Jung, Yitong Li, Siyuan Shen, Rahul~Parthasarathy
  Srikanth, Lorenzo Garattoni, Sven Meier, Nassir Navab, and Benjamin Busam.
\newblock Phocal: A multi-modal dataset for category-level object pose
  estimation with photometrically challenging objects.
\newblock In {\em Proceedings of the IEEE/CVF Conference on Computer Vision and
  Pattern Recognition}, pages 21222--21231, 2022.

\bibitem{Wang19e}
Yue Wang and Justin~M Solomon.
\newblock Deep closest point: Learning representations for point cloud
  registration.
\newblock In {\em International Conference on Computer Vision}, pages
  3523--3532, Seoul, Korea, 2019.

\bibitem{Wang19f}
Yue Wang and Justin~M Solomon.
\newblock Prnet: Self-supervised learning for partial-to-partial registration.
\newblock In {\em Advances in Neural Information Processing Systems}, pages
  8812--8824, Vancouver, British Columbia, Canada, 2019.

\bibitem{Wang18b}
Y. Wang, Y. Sun, Z. Liu, S. Sarma, M. Bronstein, and J.M. Solomon.
\newblock Dynamic graph cnn for learning on point clouds.
\newblock In {\em ACM Transactions on Graphics (TOG)}, TOG, 2019.

\bibitem{Wu15}
Zhirong Wu, Shuran Song, Aditya Khosla, Fisher Yu, Linguang Zhang, Xiaoou Tang,
  and Jianxiong Xiao.
\newblock 3d shapenets: A deep representation for volumetric shapes.
\newblock In {\em Conference on Computer Vision and Pattern Recognition}, pages
  1912--1920, Boston, MA, USA, 2015.

\bibitem{Xiang16}
Yu Xiang, Wonhui Kim, Wei Chen, Jingwei Ji, Christopher Choy, Hao Su, Roozbeh
  Mottaghi, Leonidas Guibas, and Silvio Savarese.
\newblock Objectnet3d: A large scale database for 3d object recognition.
\newblock In {\em European conference on computer vision}, pages 160--176.
  Springer, 2016.

\bibitem{Xiang18}
Yu Xiang, Tanner Schmidt, Venkatraman Narayanan, and Dieter Fox.
\newblock Posecnn: A convolutional neural network for 6d object pose estimation
  in cluttered scenes.
\newblock In {\em Robotics: Science and Systems Conference}, Pittsburgh, PA,
  USA, 2018.

\bibitem{Yang18}
Dawei Yang and Jia Deng.
\newblock Shape from shading through shape evolution.
\newblock In {\em Proceedings of the IEEE Conference on Computer Vision and
  Pattern Recognition}, pages 3781--3790, 2018.

\bibitem{Yang20a}
Dawei Yang and Jia Deng.
\newblock Learning to generate 3d training data through hybrid gradient.
\newblock In {\em Conference on Computer Vision and Pattern Recognition}, pages
  779--789, 2020.

\bibitem{Yang19}
H. Yang and L. Carlone.
\newblock A polynomial-time solution for robust registration with extreme
  outlier rates.
\newblock In {\em Robotics: Science and Systems Conference}, Freiburg im
  Breisgau, Germany, 2019.

\bibitem{Yang20}
H. Yang, J. Shi, and L. Carlone.
\newblock Teaser: Fast and certifiable point cloud registration.
\newblock In {\em arXiv Preprint}, 2020.

\bibitem{Yang15}
Jiaolong Yang, Hongdong Li, Dylan Campbell, and Yunde Jia.
\newblock Go-icp: A globally optimal solution to 3d icp point-set registration.
\newblock {\em TPAMI}, 38(11):2241--2254, 2015.

\bibitem{Yew20}
Zi~Jian Yew and Gim~Hee Lee.
\newblock Rpm-net: Robust point matching using learned features.
\newblock In {\em Conference on Computer Vision and Pattern Recognition},
  Online, 2020.

\bibitem{Yuan20}
Wentao Yuan, Benjamin Eckart, Kihwan Kim, Varun Jampani, Dieter Fox, and Jan
  Kautz.
\newblock Deepgmr: Learning latent gaussian mixture models for registration.
\newblock In {\em European Conference on Computer Vision}, pages 733--750.
  Springer, 2020.

\bibitem{Zaheer17}
Manzil Zaheer, Satwik Kottur, Siamak Ravanbakhsh, Barnabas Poczos, Russ~R
  Salakhutdinov, and Alexander~J Smola.
\newblock Deep sets.
\newblock In {\em Advances in Neural Information Processing Systems}, pages
  3391--3401, Long Beach, California, United States, 2017.

\bibitem{Zamir18}
Amir~R Zamir, Alexander Sax, William Shen, Leonidas~J Guibas, Jitendra Malik,
  and Silvio Savarese.
\newblock Taskonomy: Disentangling task transfer learning.
\newblock In {\em Proceedings of the IEEE conference on computer vision and
  pattern recognition}, pages 3712--3722, 2018.

\bibitem{Zhou16}
Qian-Yi Zhou, Jaesik Park, and Vladlen Koltun.
\newblock Fast global registration.
\newblock In {\em European Conference on Computer Vision}, pages 766--782,
  Amsterdam, the Netherlands, 2016. Springer.

\end{thebibliography}
}

\end{document}